# Predicting Regional Classification of Levantine Ivory Sculptures: A Machine Learning Approach

Amy Rebecca Gansell[1], Irene K. Tamaru[2], Aleks Jakulin[3], and Chris H. Wiggins[2]

[1]Department of the History of Art and Architecture
Harvard University
Cambridge, MA, USA
[2]Department of Applied Physics and Applied Mathematics
The Fu Foundation School of Engineering and Applied Science
Columbia University
New York, NY, USA
[3]Department of Statistics and
Institute for Social and Economic Research and Policy
Columbia University
New York, NY, USA.
gansell@fas.harvard.edu

**Abstract**

Art historians and archaeologists have long grappled with the regional classification of ancient Near Eastern ivory carvings. Based on the visual similarity of sculptures, individuals within these fields have proposed object assemblages linked to hypothesized regional production centers. Using quantitative rather than visual methods, we here approach this classification task by exploiting computational methods from machine learning currently used with success in a variety of statistical problems in science and engineering. We first construct a prediction function using 66 categorical features as inputs and regional style as output. The model assigns regional style group (RSG), with 98 percent prediction accuracy. We then rank these features by their mutual information with RSG, quantifying single-feature predictive power. Using the highest-ranking features in combination with nomographic visualization, we have found previously unknown relationships that may aid in the regional classification of these ivories and their interpretation in art historical context.

## 1 Introduction

Thousands of ivory carvings have been excavated in the Near East over the past century and a half (Thureau-Dangin 1931; Loud and Altman 1938; Barnett 1957; Hrouda 1962; Mallowan and Herrmann 1974; Herrmann 1986; Safar and al-Iraqi 1987; Herrmann 1992b). It is commonly argued that this corpus of carvings can be summarized in terms of a few general styles, believed to be associated with varying regions. While scholars have made much collective headway in the regional classification of ancient Near Eastern ivory sculpture using visual methods, it has been compellingly acknowledged that new, quantitative, and more rigorous approaches are needed (Winter 1992; Winter 1998; Winter 2005). We here respond to this need.

Most of these artifacts are derived from first millennium BC Neo-Assyrian royal contexts in northern Mesopotamia/Iraq. Here, in palaces and temples, sculptured ivories embellished the surfaces of walls, furniture, vessels and containers, and even equestrian gear. While some of the carvings fall into the Assyrian visual tradition (Herrmann 1997), the majority are unmistakably of Levantine origin.

Texts and images indicate that Neo-Assyrian rulers collected and displayed such carved ivory as tribute and booty from the Levant (Thomason 1999:393-401; Herrmann 2000:269; Herrmann and Millard 2003). Knowledge of the more specific origin of these artifacts could help define ancient economic and political dynamics. It could also be applied in the relative and absolute dating of historic first millennium BC phases and events, from which, in turn, archaeologists and art historians could extrapolate refined artifact chronologies and gain a better understanding of the temporal and geographic transmission of artistic motifs and styles (e.g., Winter 1976:20-2, 1983:185-7, 1989, 1998:150-1; Herrmann 2000:275-6). Of further cultural consequence, the elite collection and display of ivories may be interpreted to reveal ancient aesthetic appreciation and/or ideological expression (for example, Herrmann 1989; Thomason 1999; Herrmann 2000:268-9), the nuances of which may be better understood once the origins of the works are more precisely identified. Additionally, the regional classification of ivory products concerns researchers interested in ancient craft production, especially the organization of artisans and ateliers.

Archaeologists have not yet located any ancient Near Eastern ivory workshops or production centers. Nonetheless, Levantine ivories are generally attributed to visually distinct Phoenician and North Syrian carving traditions based primarily on stylistic and iconographic comparison to other art. A third possible regional classification, at times specified as an "Intermediate'" or "South Syrian" tradition, has also been considered (e.g., Winter 1981, 1998:152; Herrmann 1986:6, 1992:3; Winter Herrmann 2000:271). However, the legitimacy of this third group is now in question, and the binary labels "Phoenician" and "North Syrian" are themselves open to revision (Wicke in press).

Although one can visually organize Levantine ivories into groups of "most-Phoenician-like," "most-North Syrian-like," "most-in-between," and "anomalous," this is a subjective process, based more on art historical conditioning than on the realities of ancient craftsmanship. Ranked and clustered characteristics defining regional types and subtypes have not yet been established based on quantitative analyses of the data represented on the ivories. Instead, multivariate classification criteria are derived from the (inherently biased) visual perception of the modern beholder (Suter 1992; Winter 1998; Herrmann 2000:271-2; Winter 2005:34).

There are a number of drawbacks to relying exclusively on visual methods. Most notable is the difficulty or impossibility of revealing significant correlates among hundreds or thousands of possible pairs of features. Moreover, the ancient significance *or insignificance* of attributes and their variations (i.e., the "weight" and taxonomic "rank" of a variable) may be misinterpreted when judged outside their original cultural contexts.

Previous efforts to analyze patterns of visual and anthropometric variation in ancient Near Eastern figural art more objectively have employed cluster analysis and correspondence analysis (see, for example: Guralnick 1978; Roaf 1983; Azarpay 1990; Robins 1990). Simple ratio comparisons have been conducted on a small but coherent sample of the ivories as well (Winter 1981:105).

Due to the dimensionality of the data in our corpus, our approach favors the use of machine learning methods (for regional prediction via classification) in conjunction with information-theoretic approaches (for interpretation of single feature predictive power).

This manuscript presents our methodology and current results. We first describe the corpus, the data, and the preprocessing of data used before training the classifier. Classification performance and object analysis are followed by feature analysis performed through the use of mutual information and a nomographic visualization technique for Naïve Bayesian classification results. We conclude with future directions and a discussion of the prospective contributions of this project to the art historical field of ancient Near Eastern ivory studies.

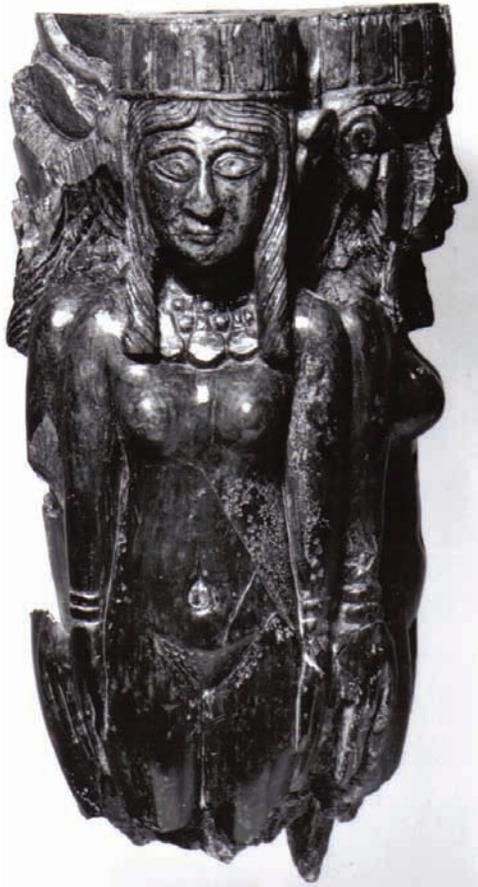

*Figure 1. Ivory furniture support with four carved female figures, from Nimrud. Ht. 9.91 cm. [The Metropolitan Museum of Art, Rogers Fund, 1952, (52.23.2) Image © The Metropolitan Museum of Art].*

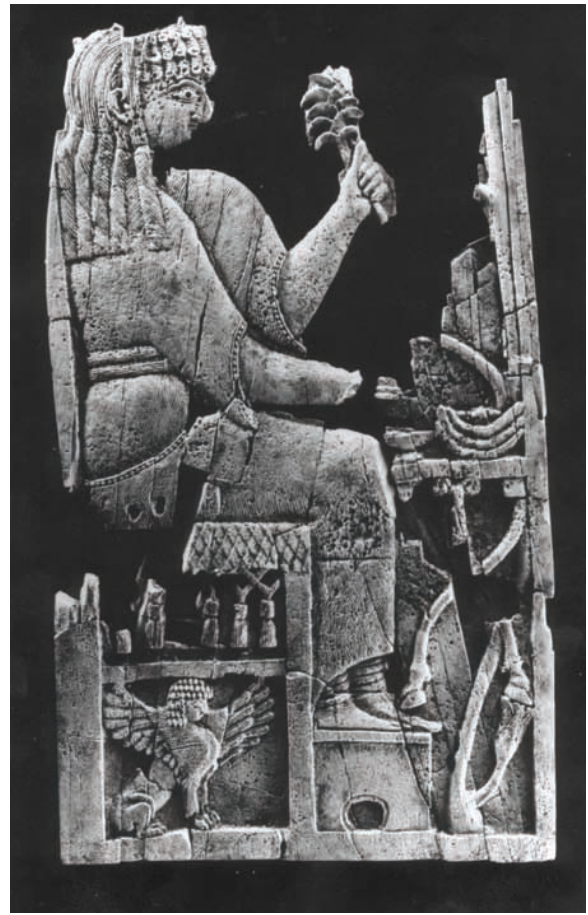

*Figure 2. Seated female at banquet, from Nimrud. Ht. 25 cm. [Iraq Museum, Baghdad, IM 60553].*

## 2 Materials, Data, and Methods

Levantine ivories, derived from the tusks of African elephants, are carved both in relief and in the round. They display a somewhat constrained iconographic repertoire, primarily depicting vegetal designs and animal and human figures. In most cases, Levantine motifs transcend regional stylistic boundaries, indicating a standard cross-regional and cross-cultural visual range. The ivories shown below have been selected to best illustrate this phenomenon. (Figures 1-8, Note that to encourage more objective consideration art historical regional designation is not provided in the captions.)

Our research focuses on images of women in particular. Female imagery is abundant and displays the greatest degree of individual variation and detail in the corpus. Still, there is only a standard assortment of female types depicted in a fixed selection of iconographic formats, including full length figures, sphinxes, and "women at the window." Full length female sculptures are portrayed nude (Figure 1), but clothed women are depicted as well, often in narrative contexts such as banquets (Figure 2) and mythological scenes. Some full length figures, both nude and clothed, have wings, and fantastical creatures such as sphinxes with female heads are commonly represented. (Figures 3, 4) Another iconographic format portrays a female head in a window frame; these works are referred to as the "woman at the window" plaques. (Figures 5, 6) Finally, many isolated and often fragmentary heads survive. (Figures 7, 8) Originally these may have belonged to full length figures, sphinxes, and women at the window. Although at first it seems easy to reconstitute heads to figural types, one must proceed carefully in assigning (let alone mathematically predicting) the type of figure a head may have belonged to, as there are several cases in which nearly identical heads have been discovered on intact sculptures of different types.

Our study sample entails 210 whole and fragmentary sculptures portraying women in all known iconographic formats. Very damaged works, heavily restored examples, and figures of ambiguous gender are not included. Nearly all relevant and reasonably accessible ivories in American, European, and Middle Eastern museum collections are represented. At the time of this research, however, hundreds of relevant ivories in Iraq were inaccessible, and their conservation conditions are unknown.

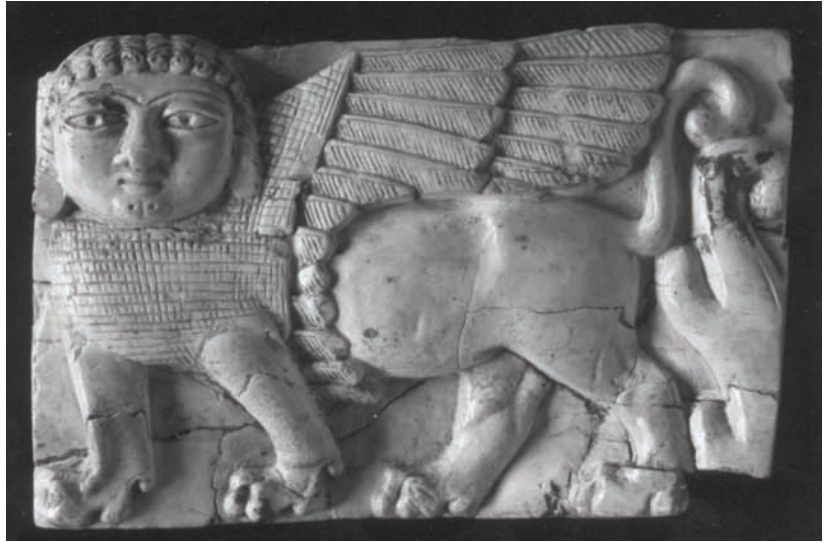

*Figure 3. Sphinx, from Nimrud. Ht. 6.6 cm. [Iraq Museum, Baghdad, IM 65280].*

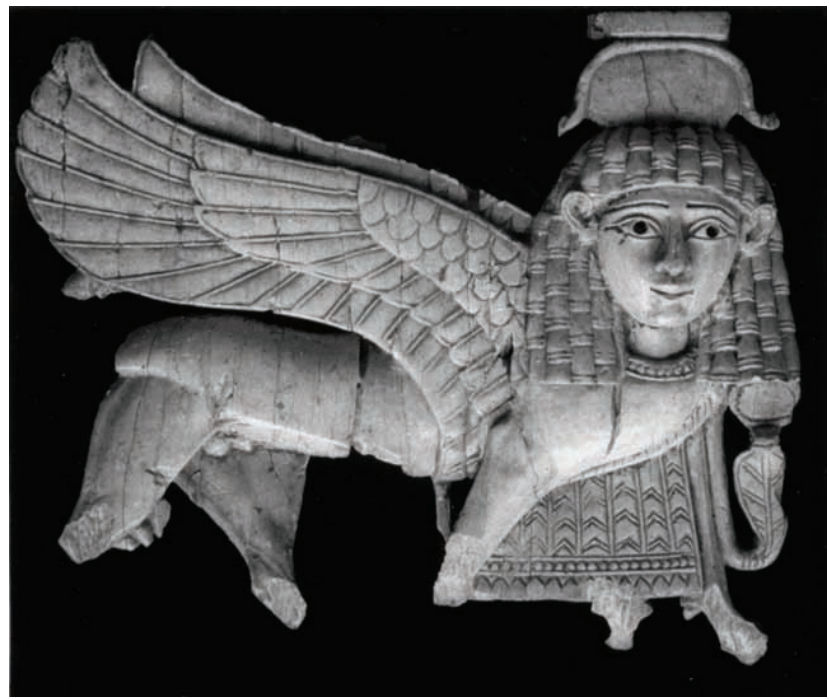

*Figure 4. Sphinx, from Nimrud. Ht. 8.79 cm. [The Metropolitan Museum of Art, Rogers Fund, 1964 (64.37.1) Image © The Metrpolitan Museum of Art].*

485

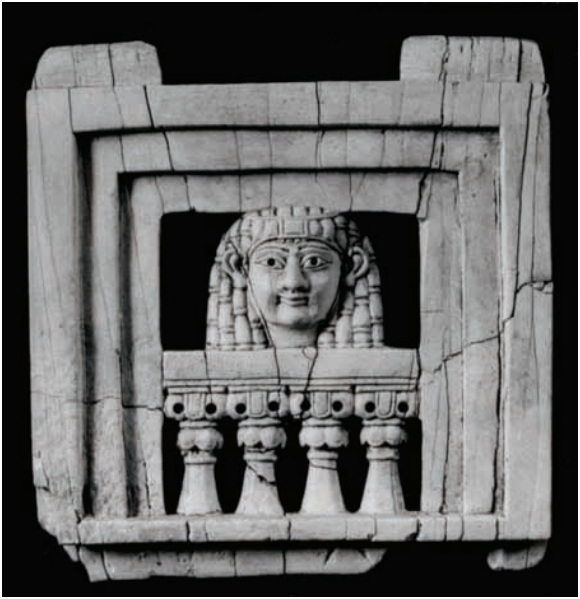
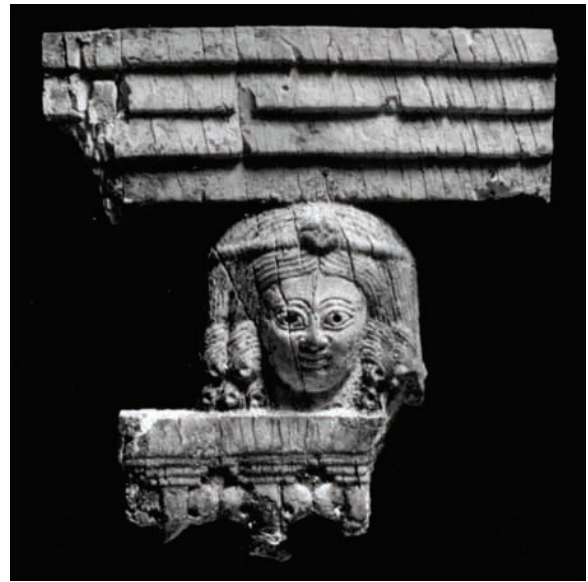

*Figure 5. Woman at the window plaque, from Nimrud. Ht. 8.2 cm. [Iraq Museum, Baghdad, IM 60500].*

*Figure 6. Woman at the window plaque, from Nimrud. Ht. 7.9 cm. [Courtesy of The British Museum, BM 8015].*

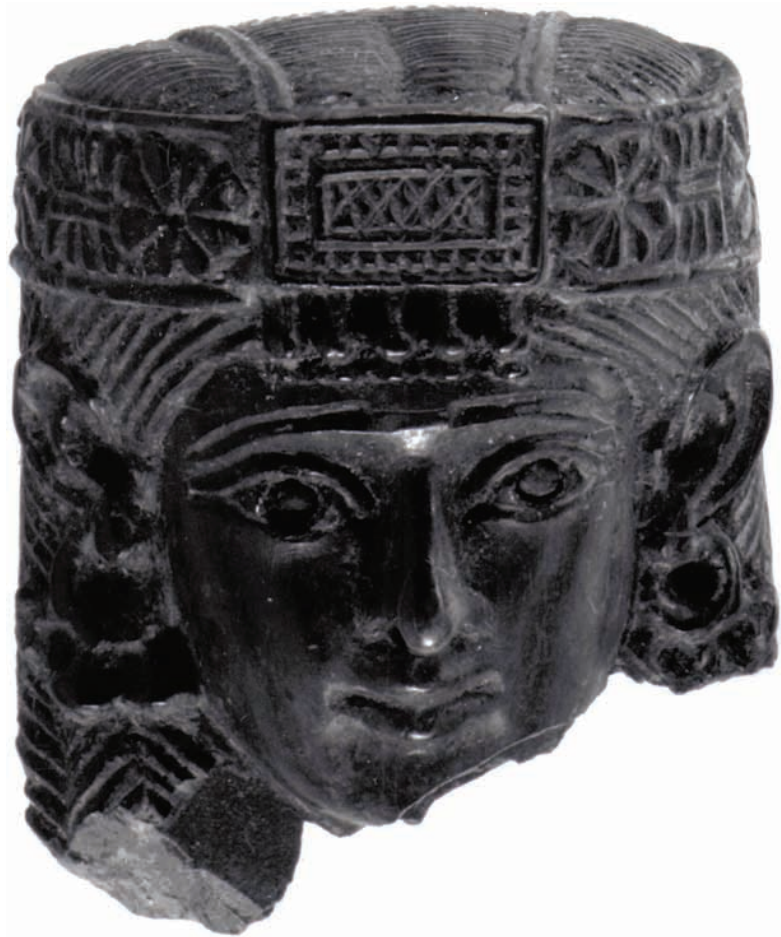

*Figure 7. Female head, from Nimrud. 4.19cm. [The Metropolitan Museum of Art, Rogers Fund, 1954 (54.117.8) Image © The Metropolitan Museum of Art].*

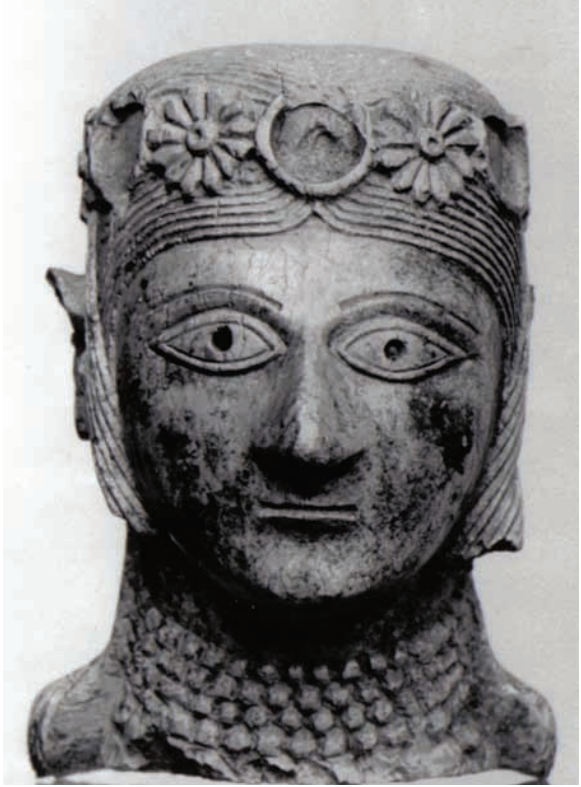

*Figure 8. Female head, from Nimrud. Ht. 4.3 cm. [Courtesy of The British Museum, BM 118234].*

### 2.1 Data

All data were collected through firsthand examination. For reference purposes, objects were drawn and photographed. A standard record of iconographic and qualitative observations was made for each ivory, and, using calipers, up to 75 point-to-point measurements were taken for each figure. The resulting dataset comprises nearly 32,000 numerical and categorical entries.

The data were input into a spreadsheet file, then saved as a comma-separated file for handling and processing by various statistical and machine learning software programs. Before processing, the data was proofread manually and edited with custom-designed UNIX-based shell scripts, employing "sed," "awk," and "grep," as well as editing tools available in Microsoft Excel. Hierarchical data were refined by creating taxonomic typologies of general characteristics (such as types of dress). Proportional ratios (such as eye height to face height, or head height to full figure height) were generated from raw measurement values and combined with the categorical data to produce a single dataset. The portion of research presented here is strictly limited to the analysis of the categorical data.

### 2.2 Methods

The principal method employed in this paper is supervised learning, a type of machine learning in which a high-dimensional (many-featured) input is used to predict an output variable (here, the categorical variable RSG). In the last decades, machine learning algorithms have advanced considerably in computational performance, efficiency, and availability. In the hopes of encouraging broader use of such statistical analyses, we feature here two of the most intuitive, user-friendly, and freely available (open-source) machine learning software suites: Weka (Witten and Frank 2005) and Orange (Demsar et al. 2004).

Weka consists of a set of machine learning algorithms which can be invoked from custom Java code or utilized directly through its graphical "Explorer mode." It may be downloaded at http://www.cs.waikato.ac.nz/ml/weka. We have thus far used a Weka-based support vector machine (SVM) classifier on the current dataset. Of the SVM implementations available in Weka, we found the Sequential Minimal Optimizer (SMO), a simple, fast algorithm which works well with sparse datasets (Platt 199:186), to be the most effective classifier of our fragmentary archaeological data. SVMs have been employed in a number of real-world applications, including handwriting recognition (Burges 1998:122-123). Aptly, unaware of what might be offered by support vector machines, a recent art historical description of the complexities involved in ivory classification referred to the challenge as "not unlike that faced by forensic experts who deal with handwriting" (Winter 2005:30).

Orange is a component-based data mining suite with similar functionality to that of Weka. Built on C++ components, it can be called via custom Python scripts or accessed through user-friendly GUI objects called Orange Widgets. It may be downloaded from http://www.ailab.si/orange/. Using single feature rankings obtained through mutual information, exploitation of Orange's nomographic visualization technique for Naïve Bayesian classification results proved effective in examining the contribution of the highest ranking features to RSG prediction (Mozina et al. 2004).

## 3 Classification Performance and Object Analysis

Statistical model-building is a worthless exercise if we cannot test the model's predictive power, i.e., whether the classifier works. The sole quantitative measure of predictive performance is the empirical estimation of the generalization error, also termed the misclassification rate, error rate, or *test loss*. To compute the test loss, we divide the objects into a training set (examples which are used to build the classifier) and the test set (examples not seen during the training). Traditionally this is done by breaking the full dataset into sets called "folds"—the training set is then all but one of these folds and the testing set is the remaining fold. For the resulting training loss and testing loss to be most meaningful, this is then averaged over the folds. Breaking the data into ten folds, as in the performance results reported in this section, is called "10-fold cross validation." We then evaluate the classification accuracy on these held-out or test sets. Ignoring this step, or merely minimizing the error on the data used to build the classifier, would be akin to merely assuming that the model works, or over-fitting the data, respectively.

Cross-validation also allows us to perform outlier detection. We do this by constructing 1,000 separate classifiers, based on different partitions of the data into test and training sets, and assigning an object-specific misclassification rate over these trials. In addition to the overall classification results, we describe below the highest-ranking of these features and suggest how these rates may be interpreted in terms of individual-object anomalies.

Using "Regional Style Group" (RSG) as class label, the SMO assigns RSGs with a testing accuracy of 98% (i.e., test loss of 2% under 10-fold cross-validation) over a set of 1,000 runs. Upon visual examination of the misclassified instances, we find it significant that the majority of these figures are outliers, that is, figures which are in some way anomalous to their labeled RSG.

For example, object BM118264, an "Intermediate/South Syrian" figure classified as "North Syrian" 918 times in 1,000 trials by the algorithm (error rate of 91.8%), is an unusually finely carved example of a winged female. Of similar figures, this is the only work of the "Intermediate/South Syrian" type to have no Egyptian-style "pegwig."

Object 65.924Bost, labeled as "North Syrian" but classified as "Intermediate/South Syrian" at an error rate of 74.9%, is a very sloppily carved "sphinx." The lack of finesse in craftsmanship distorts some of the soft tissue features which could contribute to its misclassification: it shares a bulbous chin with its "North Syrian" counterparts but lacks the double-chin which usually accompanies such a feature. Its "lip-form" as well is more prototypically "Intermediate/South Syrian."

Object BM118186, labeled as "Intermediate/South Syrian" but classified as "Phoenician" with an error rate also 74.9%, is an unusual object with no equivalent in the entire sample. The visual classification of this piece is not immediately recognizable as it represents an amalgamation of regional styles. It is the only figure of the "Intermediate/South Syrian" group with iconographic format "head" to be carved in the round, and the only figure whose eyes and eyebrows were once inlaid. The misclassification of this figure could be attributed to these particular features and to its hairline shape as well, features which are prototypically "Phoenician."

Object BM118208, of iconographic format "head," is labeled "North Syrian" but is classified as "Intermediate/SouthSyrian" with an error rate of 33.8%. Unlike 99% of its group (iconographic format "head" / RSG "North Syrian"), it is carved in relief, similar to figures of the same iconographic class in the "Intermediate/South Syrian" group. It is also the only "North Syrian" head to carry a headcharm. Though the headcharm in itself is unusual, what is of more significance is that the only other "North Syrian" figures which wear headcharms are full length figures. As such, we could hypothetically reconstitute this head to a full length figure.

Upon similar examination of the classification of unlabeled instances, we again find significance in the labels applied by the algorithm. Object BM 92233, a large-scale stone head carved in the round, was classified as "North Syrian" by the algorithm. While this sculpture is not a member of the ivory corpus, its features and proportions indicate a strong connection with the "North Syrian" RSG. Even a very fragmentary piece of a related stone sculpture (BM 139615) was classified by the machine as "North Syrian."

Far from being a repudiation of visual classification techniques, our results here highlight the accuracy and efficacy of computational techniques to augment what might be intuited by art historians and archaeologists and, thus, may be used to refine, a posteriori, certain regional classifications. For example, the assignation of "Intermediate/South Syrian" to objects labeled either "North Syrian" or "Phoenician" could suggest that "Intermediate/South Syrian" is not a true category but instead represents the overlapping edges of larger Phoenician and North Syrian groupings. Further study will need to be made to elucidate this claim.

## 4 Feature Analysis

The excellent performance in terms of prediction accuracy makes clear that there is structure to be learned within the data, but does little to elucidate which variables correlate and/or give predictive power. We now turn to the feature analysis necessary to reveal and interpret this underlying structure.

Through the use of mutual information and nomographic visualization of classifier results, we have been able to reveal previously unknown relationships between the categorical variables of the corpus.

### 4.1 Mutual Information

Mutual information (MI) can be used to find individual categorical features that offer statistically significant power for predicting a given target class (here, RSG). Since we allow the data to decide—rather than appealing to human input—we can use this approach to reveal and render previously unknown associations between RSG and individual features. (The results described in this section were obtained through the use of original MATLAB source code freely available at (http://www.artstat.sourceforge.net).

Because different features will have a different numbers of possible values, e.g., "eyeform" contains five possible "values," while "nostrilmake" (i.e., method of carving nostrils) contains only two ("excised" or "drilled"), our implementation normalizes mutual information via the measure ($\rho$), the mutual information divided by the minimum of the entropy of either $X$ or $Y$:

$$\rho = I(X; Y) / \min [H(X), H(Y)]. \qquad (1)$$

Since the amount of mutual information shared between two variables can be no larger than the minimum of the single-feature entropies, this quantity $\rho$ ranges between 0 and 1.

To illustrate, we considered the set of all (68) categorical features paired with the feature RSG. To assess the statistical significance of each pair's mutual information, we construct a background distribution by randomizing the attributes via permutation, i.e., we permuted over the corpus which objects exhibit which values for the feature being considered. This statistical test (sometimes called the "exact method") preserves the distribution of values for each feature and is a common method for assessing statistical significance (Pitman 1937).

In order to consider only those mutual information values that are statistically significant, we first restrict our attention to only those for which all of 1,000 permutations exhibit values smaller than the MI value observed for the true corpus. This leaves 22 features. Of the significant features, we restrict ourselves to the top half in terms of the normalized MI ($\rho$) as defined in Eqn. 1. This leaves 11 features that, including the counts (the number of objects for which the feature was defined) are: RegionalStyleSUBGroup (22), nostrilmake (48), posture (54), dress (80), eyeform (141), hairlineshape (145), curls (165), straight (173), pegwig (182), isIntermediate/SouthSyrian (210), isNorthSyrian (210).

From an art historical perspective, the statistical significance of categories relating to hair and eyes is noteworthy. It parallels an ancient emphasis of these features in ancient Near Eastern art; eyes and hair are consistently rendered in great detail, and often to outsized proportions. Moreover, contemporary ancient literary descriptions of beloved ones, praised rulers, and deities specify the attractiveness and allure of these features.

Because "RegionalStyleSUBGroup," "isIntermediate/SouthSyrian," and "isNorthSyrian" are obviously related to RSG, the high rankings of these categories, although reassuring in terms of validating the method, are not significant to our question. Of the remaining options, "posture," "dress," "pegwig," "straight," and "curls" are all commonly observed characteristics that have been previously used in visual classification. The remaining categories are "nostrilmake" and "eyeform." Both are of interest here because they are not generally considered by visual researchers as significant indicators of RSG.

As "nostrilmake" is only a binary category ("excised" or "drilled"), we opt to discuss immediately the more complex category of "eyeform," the entries of which include "single," "doubletop," "double," and "inlayhole." These designations refer to the articulation of eyelids on eyes carved in relief or (for "inlayhole") the presence of a hollow that would have held a composite inlaid eye (Figure 10). (The significance of "nostrilmake" and "eyeform" as a linked pair is considered below in the section on nomographic visualization.)

Graph 1, showing nearly four times as many objects with "eyeform" "double" (versus "doubletop") carry the "North Syrian" regional designation, suggests that a convention of carving eyes with distinct upper and lower lids was likely associated with what is considered North Syrian craftsmanship.

Looking at the manner in which eyes are carved on other artistic media of known North Syrian provenance, compared to those of known non-North Syrian provenance, may further strengthen the present hypothesis that "double eyeform" is a strong indicator of North Syrian production origin, or may even be part of a pure North Syrian prototype. It should be kept in mind, however, that some objects designated by art historians as "North Syrian," have eyes carved in the "doubletop" manner (i.e., both an upper and lower lids are articulated). This does not necessarily mean that these objects could not belong to the North Syrian regional carving tradition; they may simply be more peripheral (either geographically or artistically) from a "master" prototype. However, discrepancy in "eyeform" *could* indicate that objects have been misclassified by established visual methods.

## 4.2 Nomographic Visualization of Classifier Results

Orange provides a means to produce nomographic visualizations from Naïve Bayesian (NB) classification results (Mozina et al. 2004). The visualizations are particularly helpful because they allow us to see not only how much each attribute contributes to specific regional style designations, but also what exactly each value of each attribute implies. Also, by using the "attribute selection" widget in Orange, we can choose specific features to analyze further based on the information provided by MI.

Graph 2 is a nomogram of NB classification results based on the observation of "eyeform" alone. "Eyeform" "double"

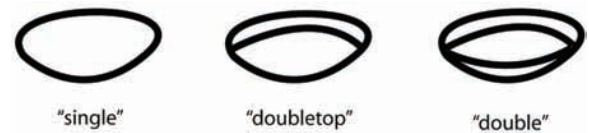

*Figure 10. Schematic illustration of carved "eyeforms."*

*Graph 1. Of all objects whose RSG is "North Syrian," there are nearly four times as many objects with "eyeform" "double" than eyeform "doubletop." The values of ($\rho$) and the mutual information are found at the top of the graph.*

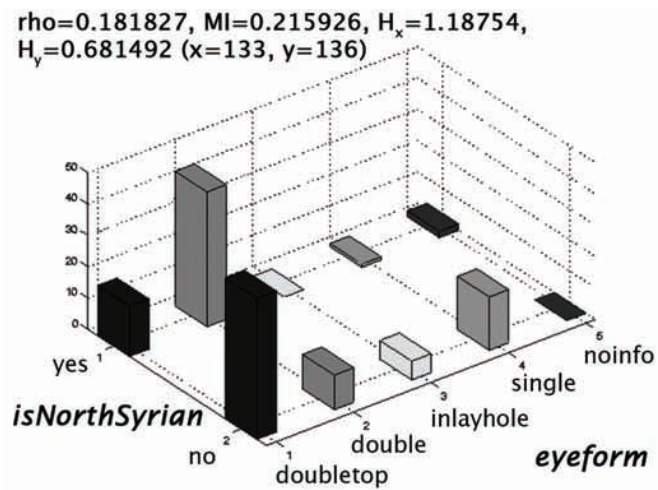

was awarded a score of 100 points out of 110 available, i.e., given an object carries "eyeform" "double," the probability of its being of North Syrian designation is 91%.

*Graph 2. Nomogram illustrating the probability of objects with "eyeform" = "double" being "North Syrian."*

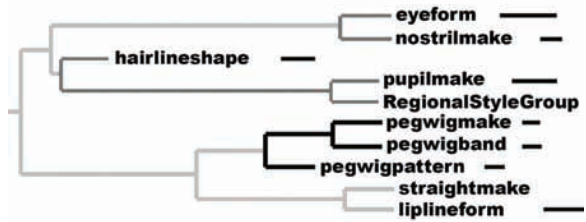

In Graph 3, the nomogram shows NB classifier results based on the observation of both "eyeform" and "nostrilmake." This tells us that when nostrils are "excised" rather than "drilled," *and* the "eyeform" is "double," the probability of the object being of North Syrian designation exceeds the 95% confidence interval mark.

These results are of particular interest as they effectively demonstrate the contribution of quantitative analysis to ivory classification. Again, the entries "eyeform" and "nostrilmake" are not generally considered by visual researchers as significant indicators of RSGs, but prove to be highly informative classification criteria.

*Graph 3. Nomogram illustrating the probability of objects with "eyeform" = "double" and "nostrilmake" = "excised" being "North Syrian."*

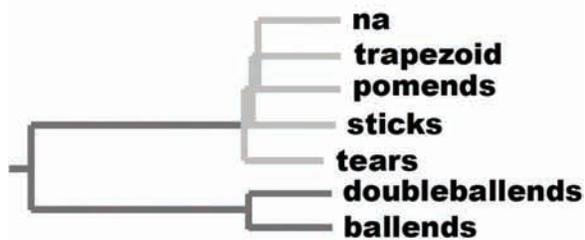

## 5 Future Outlook and Caveats

Our investigation thus far has focused solely on the categorical features contained in our dataset. There remains much to be done with the real-valued features, particularly the ratios of feature measurements (as explored in Winter 1981:105). Promising future extensions of this work include the use of algorithms that use real-valued features, along with the categorical features, as inputs.

As a specific promising direction for future statistical studies with real-valued features, we note that identifying informative features and feature-sets is the necessary prerequisite to inferring multivariate feature subsets of anthropometric, iconographic, and qualitative features that are most faithfully indicative of regional prototypical forms. These quantitatively-established regional style group prototypes could then be compared with visually-established art historical regional classifications, thereby testing conventionally-established boundaries. Based on results presented above, we expect Phoenician and North Syrian designations to remain distinct; nonetheless quantitative analysis is in progress to test this expectation and to demonstrate whether any additional distinct regional types might be revealed statistically.

By producing new schemes for the regional classification of ivory carving, we can contribute to the field of ancient Near Eastern studies in a broad sense, including by revealing historical and political dynamics. Indeed, we may even bring the modern viewer closer to understanding the ancient craftsmen's and audiences' visual and cognitive perceptions and cultural attributions of these artworks. As a corollary to this particular investigation, we hope to begin to decode conceptual templates of ancient Near Eastern feminine "beauty" (Gansell in press).

Beyond the theoretic, our final results may provide models for both physical and computer-generated artifact restoration and could potentially be applied in forgery detection. These applications are of particular relevance today, considering the conservation/restoration needs of Iraq's ivory collection and the potential dissemination of original ivory sculptures and forgeries on the antiquities market.

### 5.1 Caveats

It is useful to consider possible future contexts in which machine learning approaches, or high-dimensional approaches more broadly, might or might not be of general applicability. We note that in building a classifier—a model which takes as input high-dimensional features and as output one of a few possible class values—we require examples for which this output value is labeled. For deeper questions of art historical nature, for example, the extent to which one can quantify individual correlates with an object's "beauty," we face the daunting, but not impossible, task of obtaining such a label before commencing such an analysis.

Second, in order to deal with high-dimensional problems, we must necessarily have a large number of examples. The low generalization error presented in our results suggests that we here have a sufficiently abundant corpus of examples to reveal the underlying correlation between features and regional style groups. The "test loss" provides an empirical estimate of the generalization error and therefore the extent to which our learning problem is sufficiently sampled to reveal the underlying statistical structure. In considering additional sets of quantitative features (e.g., the myriad quantitative features such as lengths, diameters, and distances within objects), we may possibly require additional example images.

## Acknowledgements

The following institutions made this project possible by generously granting permission and access to study the ivories in their collections: the Aleppo National Museum (Syria), The Ashmolean Museum (Oxford, UK), Das Badisches Landesmuseum Karlsruhe (Germany), Birmingham Museums and Art Gallery (UK), The Boston Museum of Fine Arts (US), The British Museum (London), The Louvre (Paris), The Manchester Museum (UK), The Metropolitan Museum of Art (New York), The Museum of Archaeology and Anthropology (Cambridge, UK), and The Oriental Institute of The University of Chicago (US). Funding supporting travel and museum research was provided by a Charles Elliot Norton Dissertation Research Fellowship, the Aga Khan Foundation, and Harvard University.

This interdisciplinary collaboration was originally envisioned and encouraged by Alicia Walker, to whom we express our gratitude. Thanks also to Jake Hofman for his invaluable help with and expertise in Weka. In addition, the pioneering scholarly vigor and involvement in this project of Irene J. Winter has greatly motivated and sustained our endeavor.
## References Cited

Azarpay, Guitty. 1990. A Photogrammetric study of three Gudea statues. *Journal of the American Oriental Society* 110:660-5.

Barnett, Richard D. 1957. *A Catalogue of the Nimrud Ivories with Other Examples of Ancient Near Eastern Ivories in the British Museum*. London: Trustees of the British Museum.

Burges, Christopher J. C. 1998. A Tutorial on support vector machines for pattern recognition. *Data Mining and Knowledge Discovery* 2:121-67.

Demsar, Janez, Zupan, Blaz, Leban, Gregor. 2004. Orange: From experimental machine learning to interactive data mining. White Paper (www.ailab.si/orange). Faculty of Computer and Information Science, University of Ljubljana, Slovenia.

Gansell, Amy Rebecca. In press. Measuring beauty: An anthropometric methodology for the assessment of ideal feminine beauty as embodied in first millennium BCE ivory carvings. In, *Syrian and Phoenician Ivories of the First Millennium BCE*. Claudia E. Suter and Christoph Uehlinger, eds. Orbis Biblicus et Orientalis.

Guralnick, Eleanor. 1976. The proportions of some archaic Greek sculptured figures: A computer analysis. *Computers and the Humanities* 10:153-69.

Herrmann, Georgina. 1986. *Ivories from Room SW 37 Fort Shalmaneser*. Ivories from Nimrud IV. London: The British School of Archaeology in Iraq.

Herrmann, Georgina. 1989. The Nimrud ivories, 1: The Flame and Frond School. *Iraq* 51:85-109.

Herrmann, Georgina. 1992a. The Nimrud ivories, 2: A survey of the traditions. In, *Von Uruk nach Tuttul, eine Festschrift für Eva Strommenger, Studien und Aufsätze von Kollegen und Freunden*. B. Hrouda, S. Kroll, and P. Z. Spanos, eds., pp. 65-79. Munich: Münchener Vorderasiatische Studien.

Herrmann, Georgina. 1992b. *The Small Collections from Fort Shalmaneser*. Ivories from Nimrud V. London: The British School of Archaeology in Iraq.

Herrmann, Georgina. 1997. The Nimrud Ivories 3: The Assyrian tradition. In, *Assyrien im Wandel der Zeiten*. H. Waetzoldt and H. Hauptmann, eds., pp. 285-90. Heidelberg: Heidelberger Studien zum Alten Orient 6.

Herrmann, Georgina. 2000. Ivory carving of first millennium workshops, traditions and diffusion. In, *Images as Media: Sources for the Cultural History of the Near East and the Eastern Mediterranean (1st Millennium BCE)*. Christoph Uehlinger, ed., pp. 267-82. Orbis Biblicus et Orientalis 175. Fribourg: University Press.

Herrmann, Georgina and Millard, Alan. 2003. Who used ivories in the early first millennium BC? In, *Culture through Objects, Ancient Near Eastern Studies in Honour of P. R. S. Moorey*. T. Potts, M. Roaf, and D. Stein, eds., pp. 377-402. Oxford: Griffith Institute.

Hrouda, Barthel. 1962. *Tell Halaf IV, Die Kleinfunde aus Historischer Zeit*. Berlin: Walter de Gruyter & Co.

Loud, Gordon and Altman, Charles B. 1938. *Khorsabad II: The Citadel and the Town*. Oriental Institute Publications 40. Chicago: University of Chicago Press.

Mallowan, Max and Herrmann, Georgina. 1974. *Furniture from SW.7 Fort Shalmaneser*. Ivories from Nimrud III. London: The British School of Archaeology in Iraq.

Mozina, M., Demsar, J., Kattan, M., Zupan, B. 2004. Nomograms for visualization of Naïve Bayesian Classifier. In, *Proceedings of the 8th European Conference on Principles and Practice of Knowledge Discovery in Databases, Pisa, Italy*. Jean-François Boulicaut, et al., eds., pp. 337-48. New York: Springer.

Oates, Joan and Oates, David. 2001. *Nimrud, An Assyrian Imperial City Revealed*. London: The British School of Archaeology in Iraq.

Pitman, Edwin J. G. 1937. Significance tests which may be applied to samples from any population (Parts I and II). *Royal Statistical Society Supplement* 4:119-30 and 225-32.